\documentclass[acmsmall]{acmart}
\AtBeginDocument{%
  \providecommand\BibTeX{{%
    \normalfont B\kern-0.5em{\scshape i\kern-0.25em b}\kern-0.8em\TeX}}}

\usepackage{xspace}
\usepackage{cleveref}

\usepackage{multirow}

\newcommand{\ACL}{224\xspace}
\newcommand{\XX}{304\xspace}

\acmJournal{JACM}

\begin{document}

\title{A Survey on Gender Bias in Natural Language Processing}

\author{Karolina Sta\'nczak}
\email{ks@di.ku.dk}
\orcid{0000-0001-7326-9594}
\affiliation{%
  \institution{University of Copenhagen}
}

\author{Isabelle Augenstein}
\email{augenstein@di.ku.dk}
\orcid{}
\affiliation{%
  \institution{University of Copenhagen}
}

\renewcommand{\shortauthors}{Sta\'nczak, et al.}

\begin{abstract}
Language can be used as a means of reproducing and enforcing harmful stereotypes and biases and has been analysed as such in numerous research.
In this paper, we present a survey of \XX papers on gender bias in natural language processing. We analyse definitions of gender and its categories within social sciences and connect them to formal definitions of gender bias in NLP research. We survey lexica and datasets applied in research on gender bias and then compare and contrast approaches to detecting and mitigating gender bias. We find that research on gender bias suffers from four core limitations. 1) Most research treats gender as a binary variable neglecting its fluidity and continuity. 2) Most of the work has been conducted in monolingual setups for English or other high-resource languages. 
3) Despite a myriad of papers on gender bias in NLP methods, we find that most of the newly developed algorithms do not test their models for bias and disregard possible ethical considerations of their work. 4) Finally, methodologies developed in this line of research are fundamentally flawed covering very limited definitions of gender bias and lacking evaluation baselines and pipelines.
We see overcoming these limitations as a necessary development in future research.   
\end{abstract}



\maketitle

\section{Introduction}






Gender bias and sexism are explicitly expressed in language and thus, have been analysed both by the linguistics and natural language processing (NLP) communities \citep{sun-etal-2019-mitigating,koolen-van-cranenburgh-2017-stereotypes}. Since the first publication on gender bias detection in 2004 in the ACL Anthology\footnote{\url{https://aclanthology.org/}}, which indexes papers published at almost all NLP venues, there have been a total of \ACL publications aiming an investigation of gender bias, showing a clear upward trend in the number of papers published every year that has started back in 2015.
In particular, previous research has confirmed gender bias to be prevalent in literature \citep{hoyle-etal-2019-unsupervised}, news \citep{wevers-2019-using}, media \citep{asr2021}, and communication about and directed towards people of different genders \citep{fast2016shirtless,voigt-etal-2018-rtgender}. Further, prior studies have shown bias in underlying NLP algorithms such as word embeddings \citep{bolukbasi2016man} and language models \citep{nadeem2020stereoset}, as well as in the downstream tasks they are employed for, e.g., machine translation \citep{savoldi2021gender}, coreference resolution \citep{zhao-etal-2018-gender, rudinger-etal-2018-gender, webster-etal-2018-mind}, language generation \citep{sheng-etal-2020-towards}, and part-of-speech tagging and parsing \citep{garimella-etal-2019-womens}. 

However, the rapid increase in research on gender bias has led to a state where the research is fractured across communities and publications often do not engage with parallel research. Thus, there is a need to summarise and critically analyse the developments hitherto, to identify the limitations of prior work and suggest recommendations for future progress. Therefore, in this paper, we present an overview of \XX papers on gender bias in natural language processing. 
We begin with a brief outline our methodology and explore the evolution of the field in popular NLP venues (\S\ref{sec:method}). Then, we discuss different definitions of 
gender in society (\S\ref{sec:gender}). Further, we define gender bias and sexism in general and in NLP, in particular, incorporating a discussion of their ethical considerations (\S\ref{sec:bias}). Next, we gather common lexica and datasets curated for research on gender bias (\S\ref{sec:resources}). Subsequently, we discuss formal definitions of gender bias (\S\ref{sec:definition}). Then, we discuss methods developed for gender bias detection (\S\ref{sec:detection}) and mitigation (\S\ref{sec:mitigation}). 

We find that existing research on gender bias has four main limitations and see addressing these limitations as necessary future focus areas of research on gender bias. 
Firstly, despite the wide range of research across multiple language tasks predominantly only two genders are distinguished, male and female, neglecting the fluidity and continuity of gender as a variable. Natural language has started to adopt gender-neutral linguistic forms to recognise non-binary nature of gender such as singular \textit{they} in English and \textit{hen} in Swedish, thus presenting a need for NLP researchers to incorporate this social development into their datasets and algorithms \citep{Sun2021TheyTT}. 
Otherwise, modelling gender as a binary variable can lead to a number of harms such as misgendering and erasure via invalidation or obscuring of non-binary gender identities \citep{fast2016shirtless, behm2008}. 
Addressing this issue is critical not just to improve the quality of our systems, but more importantly to minimise these harms \citep{larson-2017-gender}. 

Secondly, most prior research on gender bias has been monolingual, focusing predominantly on English or a small number of further high-resource languages such as Chinese \citep{liang-etal-2020-monolingual} and Spanish \citep{zhao-etal-2020-gender}. Only limited work has been conducted in a broader multilingual context with notable exceptions of analysis of gender bias in machine translation \citep{prates2019assessing} and language models \citep{stanczak2021quantifying}.   

Thirdly, despite a plethora of studies showing evidence of presence of systematic gender bias in prolifically applied NLP methods \citep{bolukbasi2016man, nangia-etal-2020-crows,nadeem2020stereoset}, researchers are not required to test the models they publish with respect to biases they perpetuate. In particular, still most of the recently published models do not include a study of (gender) bias and ethical considerations alongside their publication \citep{devlin-etal-2019-bert,raffel2020exploring,conneau-etal-2020-unsupervised,zhang2020cpm} with the noteworthy exclusion of GPT-3 \citep{brown2020language}. In general, these methods are tested for biases only post-hoc when already being deployed in real-life applications potentially posing harm to different social groups \citep{mitchell2019}.  

Lastly, we argue that methodologies within gender bias detection often lack baselines and do not engage with parallel research. We find that similarly to research within societal biases \citet{blodgett2020language}, work on gender bias in particular, is fundamentally flawed suffering incoherence in usage of evaluation metrics. Publications consider often limited definitions of bias that address only one of many ways gender bias manifests itself in language.

\section{Methodology}
\label{sec:method}

The following survey is an overview of all papers identified by the authors on analysing gender bias in NLP, which spans a total of \XX papers. To collect these relevant papers, the ACL Anthology, NeurIPS, and FAccT 
were queried for all papers with the keywords `gender bias', `gender' or `bias' made available prior to June 2021. Additionally, we expand the spectrum of the papers with relevant social science publications and other relevant publications cited in the collected papers.  

We retained all papers about gender bias and discarded papers focusing on other definitions of the keywords (e.g., inductive bias, social bias). We review papers analysing gender bias in natural language and methods presenting an encompassing overview of gender bias in language. 

\begin{figure}[h]
    \centering
    \includegraphics[width=\columnwidth]{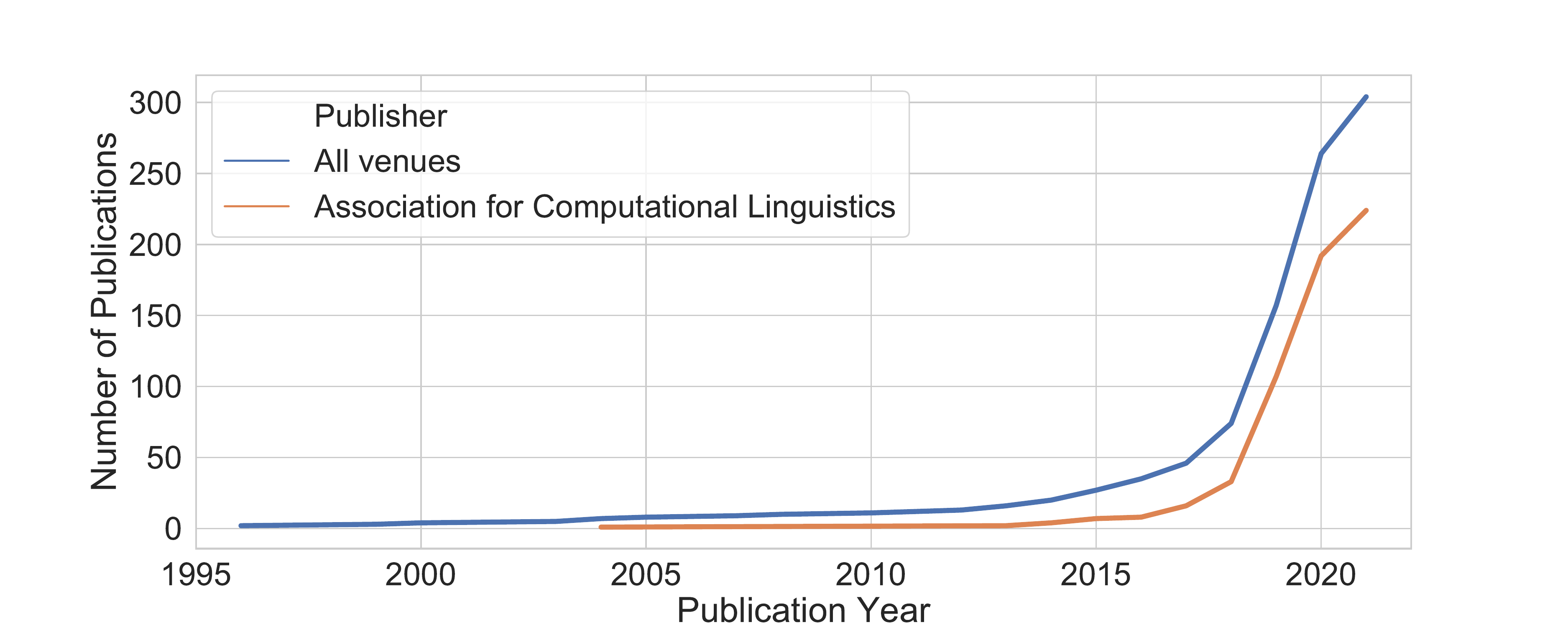}
    \caption{Cumulative number of papers published on gender bias prior to June 2021.
    }
    \label{fig:timeline}
\end{figure}

We analyse the number of published papers in ACL venues mentioning the selected keywords either in the title or the abstract of the paper and present the results in Figure \ref{fig:timeline}. We observe a steady increase in the number of papers since 2015 with notable peaks in 2019 (83 publications) and 2020 (a total of 107 publications).
This trend suggests 2021 might end with another record in the number of papers on gender bias per year. Indeed, in 2021, we have already identified a total of 40 papers covering the topic of gender bias in NLP.     
This development demonstrates that the area of research has established itself within NLP research. 

\section{Gender in Society and Linguistics}
\label{sec:gender}


Definitions of gender used in the linguistics literature vary substantially across subfields and are often implicit  \citep{ackerman2019}. Depending on the context, the concept of gender refers to a person's self-determined identity and the way they express it, how they are perceived, and others' social expectations of them \citep{ackerman2019, lucy-bamman-2021-gender}. Compared to \textit{sex}, a term that solely refers to one's set of physical and physiological characteristics such as chromosomes, gene expressions, and genitalia, \textit{gender} is considered a social construct \citep{risman2004, Butler1989-BUTGTF-2}. In particular, \citet{risman2004} argue gender is a social construct and, as such, has consequences on person's individual development, both in interactions and institutional domains. 



However, linguistic categories of gender do not map well to social categories \citep{cao-daume-iii-2020-toward}.
Literature on gender in linguistics often distinguishes the following types of gender that are summarised below. We note that these types are not all-encompassing and merely outline gender categories presented in the literature.

\begin{itemize}
    \item \textbf{Grammatical gender}: refers to a classification of nouns based on a principle of a grammatical agreement into categories. Depending on the language, the number of grammatical gender classes ranges from two (e.g., \textit{masculine} and \textit{feminine} in French, Hindi, and Latvian) to several tens (in Bantu languages and Tuyuca) \citep{corbett_1991}. Many of these languages also assign grammatical gender to inanimate nouns.
    \item \textbf{Referential gender}: identifies referents as \textit{female}, \textit{male} or \textit{neuter} \citep{cao-daume-iii-2020-toward}. A very similar concept is described by conceptual gender referred to as a gender that is expressed, inferred and used by a perceiver to classify a referent \citep{cao-daume-iii-2020-toward}.
    \item \textbf{Lexical gender}: refers to an existence of lexical units carrying the property of gender, male- or female-specific words such as \textit{father} and \textit{waitress} \citep{FuertesOlivera2007ACV,cao-daume-iii-2020-toward}.
    \item \textbf{(Bio-)social gender}: refers to the imposition of gender roles or traits based on phenotype, social and cultural norms, gender expression, and identity (such as gender roles) \citep{kramarae1985-KRAAFD, ackerman2019}.
\end{itemize}

\paragraph{Non-binary gender}

Since the grammatical, referential, and lexical gender are definitions widely followed in NLP research, most NLP research that includes gender as a variable in downstream tasks treats it as a categorical variable with binary values (in English) \citep{brooke-2019-condescending}. However, the binarisation of gender in computational studies usually does not agree with critical theorists. For instance, \citet{Butler1989-BUTGTF-2} show how gender is not simply a biological given, nor a valid dichotomy, and even though many people fit into the binary categories, there are more than two genders \citep{bing_question_1998}. Thus, gender can be viewed as a broad spectrum. 


More recently, natural language started adopting linguistic forms to recognise the non-binary nature of gender, such as singular \textit{they} in English, \textit{hen} in Swedish and \textit{h\"an} in Finnish. These linguistic forms are not new concepts and were used by native speakers to refer to someone whose gender is unknown. However, their popularity has increased to denote a person whose gender is non-binary. The increased popularity of gender-neutral linguistic forms in natural language presents a challenge to incorporate this social development into the datasets and algorithms \citep{Sun2021TheyTT}. However, some words that are relevant in this discussion such as \textit{cisgender} and \textit{binarism} are either missing or underrepresented in corpora and databases \citep{hicks-etal-2016-analysis}. 

\paragraph{Determining gender}

To include gender as a variable in a NLP method, it often needs to be determined from the data first since it is often not explicitly given which is generally difficult to accomplish with high precision. 

A popular method to determine gender is to infer it from a person's name, assuming that this information is given. In many languages, gender-differentiated names for men and women make gender assignment possible based on gendered name dictionaries. For instance, in Slavic languages, the ending of the last name is gender-specific (e.g., with -i vs. -a). 
On the other hand, gender-neutral first names are common for Chinese, Turkish, and many other languages. Additionally, names often have different gender associations depending on the country and language, such as \textit{Andrea} being a male name in Italian and a female one in English, German and Spanish. Notably, many of the primarily Western-based name lists used for determining gender do not always generalise to names from other countries and cultures \citep{lucy-bamman-2021-gender}. Due to these aspects, all of the above methods of determining gender tend to be imprecise and neglect non-binary genders.  







\section{Gender Bias, Sexism and Harms they make}
\label{sec:bias}

In the following, we state general definitions of gender bias and sexism and distinguish among their different types. Further, we outline the potential harms they might cause for individuals and society as a whole.


\subsection{Gender Bias}

\citet{blodgett2020language} warn that papers about NLP systems developed for the same task often conceptualise bias differently. Therefore, we state the most common definitions of gender bias in the following. 
Gender bias is defined as the systematic, unequal treatment based on one's gender \citep{sun-etal-2019-mitigating}. More specifically, \citet{batyanissenbaum1996bias} use the term bias to refer to behaviour that systematically discriminates against specific individuals or groups in favor of others and distinguish three bias categories: pre-existing bias, technical bias, and emergent bias. \textbf{Pre-existing bias} arises when computer systems incorporate biases that appear independently and often prior to the creation of the system \citep{batyanissenbaum1996bias}. It can originate both from individuals, biased software developers or from society, private or public organizations and institutions, or especially in case of gender bias -- historical and cultural context. Thus, this type of bias emerges not only through conscious decisions of individuals or institutions but can also appear unintended. On the other hand, \textbf{technical bias} emerges from models' technical deign such as hardware and software limitations. While it is almost always possible to identify pre-existing bias and technical bias in a system design at the time of creation or implementation, \textbf{emergent bias} arises when the context the system was used for has changed - due to changes in society, population, or cultural values (e.g., when social media feeds are influenced by user's gender).




Further literature outlines reporting and interpretation bias.
\textbf{Reporting bias} refers to the phenomenon that the frequency with which situations of a certain type are described in text does not necessarily correspond to their relative likelihood in the world, or the subjective frequency captured in human beliefs \citep{gordon2013reporting}. 
On the other hand, \textbf{interpretation bias} is a phenomenon of researchers assuming that gender is a relevant variable which ultimately leads to analyses that are incapable of revealing violations of this assumption \citep{Bamman2012GenderIA, koolen-van-cranenburgh-2017-stereotypes}. The results are not questioned, especially if they align with common often stereotypical knowledge \citep{koolen-van-cranenburgh-2017-stereotypes}. 

\subsection{Sexism}

Sexism can be defined as discrimination, stereotyping, or prejudice based on one's sex (as opposed to one's gender). According to the ambivalent sexism theory \citep{glick1996sexism}, sexism can be divided as:

\begin{itemize}
\item Hostile: follows the classic definition of prejudice - an explicitly negative sentiment that is sexist. 
\item Benevolent: subjectively positive attitude, which is sexist. Despite the seemingly positive sentiment, benevolent sexism has been shown to affect women's cognitive performance stronger than hostile sexism \citep{Dardenne2007InsidiousDO}. For instance, female gender associations with any word, even a subjectively positive word such as \textit{attractive}, can cause discrimination against women if it reduces their association with other words, such as \textit{professional}. Despite the positive sentiment of benevolent sexism, it can be backtracked to masculine dominance and stereotyping. 
\end{itemize}


We note that sexism is considered a subset of hate speech \citep{waseem-hovy-2016-hateful} and therefore is often analysed together with other forms of aggression \citep{safi-samghabadi-etal-2020-aggression}.

\subsection{Harms}

Gender bias and sexism result in harms affecting individuals and society as a whole. Recently, \citet{crawford2017keynote} present a framework classifying algorithmic biases by the type of harm they cause and distinguish between allocational and representational harms.

\textbf{Representational harms} refer to portrayals of certain groups that are discriminatory. In general, following \citet{crawford2017keynote} representational harms can be divided into:
stereotyping, under-representation, denigration, recognition, and ex-nomination. Stereotyping, in particular, perpetuates common (often negative) depictions of a certain gender.
Under-representation bias is the disproportionately low representation of a specific group. Denigration refers to the use of culturally or historically derogatory terms, while recognition bias involves a given algorithm's inaccuracy in recognition tasks. Finally, ex-nomination describes a practice where a specific category or way of being is framed as the norm by not giving it a name or not specifying it as a category in itself (e.g., `politician' vs. `female politician').
On the other hand, \textbf{allocational harms} refer to the unjust distribution of opportunities and resources due to algorithmic intervention. They can result in systematic differences in treatment or denial of a particular service and complete ruling out of certain groups, for instance in job applications. Allocation bias can be framed as an economic issue in which a system unfairly allocates resources to certain groups over others, while representation bias occurs when systems detract from the social identity and representation of certain groups \citep{crawford2017keynote, sun-etal-2019-mitigating}.

Another harmful outcome of gender bias and sexism presents itself in \textbf{gender gaps} that arise from these asymmetrical valuations, e.g., where men are typically over-represented and have higher salaries compared to women \citep{mitra-2003-gender}. The public sphere is often associated with male and agents characteristics (assertiveness, competitiveness) in domains like politics and entrepreneurship. Private or domestic domains linked to family and social relationships are traditionally related to women, although social relationships are considered more important by people independent of gender \citep{friedman-etal-2019-relating}. 



\subsection{Bias in NLP}
Above we have introduced gender bias and sexism as general terms. In the following, we discuss how these biases emerge in natural language and ultimately influence many downstream tasks. 


Language can be used as a substantial means of expressing gender bias.
Gender biases are translated from source data to existing algorithms that may reflect and amplify existing cultural prejudices and inequalities by replicating human behavior and perpetuating bias \citep{sweeney2013discrimination}. This phenomenon is not unique to NLP, but the lure of making general claims with big data, coupled with NLP's semblance of objectivity, makes it a particularly pressing topic for the discipline \citep{koolen-van-cranenburgh-2017-stereotypes}.

Alongside the types of biases described above, there are forms of bias that apply specifically in NLP research. In particular, \citet{hitti-etal-2019-proposed} define gender bias in a text as the use of words or syntactic constructs that connote or imply an inclination or prejudice against one gender. Further, \citet{hitti-etal-2019-proposed} note that gender bias can manifest itself structurally, contextually or in both of these forms. \textbf{Structural bias} arises when the construction of sentences shows patterns that are closely tied to the presence of gender bias. It encompasses gender generalisation (i.e., when a gender-neutral term is assumed to refer to a specific gender-based on some (stereotypical) assumptions) and explicit labeling of sex. On the other hand, \textbf{contextual bias} manifests itself in a tone, the words used, or the context of a sentence. Unlike structural bias, this type of bias cannot be observed through grammatical structure but requires contextual background information and human perception. Contextual bias can be divided into societal stereotypes (which showcase traditional gender roles that reflect social norms) and behavioral stereotypes (attributes and traits used to describe a specific person or gender). Therefore, gender bias can be detected using both linguistic and extra-linguistic cues, and can manifest itself with different intensities, which can be subtle or explicit, posing a challenge in this line of research.

Gender bias is known to perpetuate to models and downstream tasks posing harm for the end-users \citep{bolukbasi2016man}. These harms can emerge as representational and allocational harms and gender gaps. 
\textbf{Allocation harm} is reflected when models often perform better on data associated with the majority gender. In the context of NLP, this is often the case for machine translation \citep{sap-etal-2017-connotation} and coreference resolution \citep{webster-etal-2018-mind} (see \S \ref{sec:detect-tasks}). 
\textbf{Representation harm} is reflected when associations between gender with certain concepts are captured in word embeddings and model parameters \citep{sun-etal-2019-mitigating}, for instance, as shown in \citep{bolukbasi2016man, zhao-etal-2018-learning}. On the other hand, \textbf{gender gap} is a phenomenon influencing gender bias in the text. Since women are underrepresented in most areas of society, it is not surprising that available texts mainly discuss and quote men \citep{asr2021}, which leads, for example, to biased corpora researchers train their models on. 

\section{Resources}
\label{sec:resources}

Comprehensive data resources are crucial in probing for gender bias in language. However, many of the datasets in NLP are inadequate for measuring gender bias since they are often severely gender imbalanced with a substantial under-representation of female and non-binary instances. Further, analysing gender bias often requires a dataset of a specific structure or including certain information to enable proper isolation of the effect of gender \citep{sun-etal-2019-mitigating}. Thus, evaluation on widely-used datasets (e.g., SNLI \citep{rudinger-etal-2017-social}) might not reveal gender bias due to inherent biases encoded in the data, presenting a need in research for targeted datasets for gender bias detection. 



We note that the choice of a dataset is dependent on the considered definition of bias (discussed in \S \ref{sec:bias}) that needs to be targeted specifically, the NLP task at hand, domain, etc. Here, we describe the most popular publicly available lexica (\S \ref{sec:lexica}) and datasets (\S \ref{sec:datasets}) that have been used to analyse gender bias in NLP with respect to the above-mentioned aspects. 


\subsection{Gender lexica}
\label{sec:lexica}

Lexicon matching is an interpretable and technically simple approach, and thus, it has been frequently adopted by NLP practitioners. In particular, in gender bias detection, lexica representing genderness, sentiment, and the affect dimensions of valence, arousal, and dominance have been widely employed since these measures are often used as proxies for bias. In Table \ref{tab:lexica}, we present the most popular lexica used for gender bias detection, and in the following, we describe measures they quantify. 

\begin{table}[h]
\centering
\resizebox{\textwidth}{!}{
\begin{tabular}{lrr}
\toprule
Lexicon & No. of words & Measure
\\  \midrule
Gender Ladeness Lexicon \citep{ramakrishna-etal-2015-quantitative} & 10 000 & Genderness \\
Gender Predictive Lexicon \citep{sap-etal-2014-developing} & 7 136 & Genderness \\
Gender Ladeness Lexicon \citep{clark_paivio_2004} & 925 & Genderness \\
Williams and Best \citep{Williams1990SexAP} & 300 & Genderness \\
NRC VAD Lexicon \citep{mohammad-2018-obtaining} & 20 000 &  Valence, Arousal and Dominance \\
Valence, Arousal, and Dominance \citep{warriner-2013-valence} & 13 915 & Valence and Dominance \\
NRC Emotion Lexicon \citep{mohammad2013crowdsourcing} & 10 170 & Emotion and Sentiment \\
Connotation Frames \citep{sap-etal-2017-connotation} & 2 155 & Power and Agency \\
\bottomrule
\end{tabular}
}
\caption{List of popular lexica used in gender bias research.}
\label{tab:lexica}
\vspace*{-0.8cm}
\end{table}

\subsubsection{Sentiment} 
Differences in sentiment towards people of different genders have been analysed in the context of gender bias in numerous papers \citep{hoyle-etal-2019-unsupervised,touileb-etal-2020-gender,cho-etal-2019-measuring,stanczak2021quantifying}, which have exploited sentiment lexica for this purpose. Since creating a comprehensive overview of sentiment lexica is outside the scope of this paper, we refer the reader to \citet{taboada-2011-lexicon} for such an overview. However, we note that sentiment is indicative solely of hostile biases rather than more nuanced ones. 

\subsubsection{Gender Ladenness}
Gender ladenness is a measure to quantitatively represent a normative rating of the perceived feminine or masculine association of a word \citep{Paivio1968ConcretenessIA}. In particular, this metric indicates the gender specificity of individual words, with extreme values assigned to highly stereotypical concepts. For instance, in \citet{ramakrishna-etal-2015-quantitative}'s lexicon, which is based on movie scripts, the word \textit{bride} would be assigned the gender ladenness value of 0.84 on a scale from -1 (most masculine) to 1 (most feminine). Similarly, \citet{Williams1990SexAP} use a list of pre-selected adjectives, \citet{sap-etal-2014-developing} use words collected on social media, and \citet{clark_paivio_2004} select a list of nouns to create a genderness lexicon. 



\subsubsection{Valence, Arousal, and Dominance}
Based on social psychology, NLP research has identified three primary affect dimensions: power/dominance (strength/weakness), valence (goodness/badness), and agency/arousal (activeness/passiveness of an identity) \citep{field-tsvetkov-2019-entity}. 
Since a common stereotype associates female gender with weakness, passiveness, and submissiveness, lexica reporting measures for these dimensions are a valuable resource in gender bias analysis, and going beyond sentiment, they can be applied in unveiling benevolent biases. 


\subsubsection{Limitations}
By their nature, lexicon approaches are limited to known words \citep{field2019contextual}, and they assume that the context of the words remains constant \citep{li2020content}. However, collecting exhaustive lexica can be very resource-consuming since they rely on human-generated annotations \citep{li2020content}. Moreover, we note that all the lexica listed in Table \ref{tab:lexica} are created solely for English. There has been very little research enabling multi-lingual gender bias analysis employing lexica, with the notable exception of \citet{stanczak2021quantifying}.


\subsection{Datasets}
\label{sec:datasets}

\begin{table}[h]
\centering
\resizebox{\textwidth}{!}{
\begin{tabular}{lrrrrr}
\toprule
Dataset & Size & Data & Gender & Task & Bias \\  \midrule
EEC \citep{kiritchenko2018examining} & 8 640 sent. & sent. templates & b & SA & stereotyping \\
WinoBias \citep{zhao-etal-2018-gender}  & 3 160 sent. & sent. templates & nb & cor. res. & occ. bias \\
WinoGender \citep{rudinger-etal-2018-gender} & 720 sent. & sent. templates & b & cor. res. & occ. bias \\
WinoMT \citep{stanovsky-etal-2019-evaluating}  & 3 888 sent. & sent. templates & b & MT & occ. bias \\
Occupations Test \citep{escude-font-costa-jussa-2019-equalizing} & 2 000 sent. & sent. templates & b & MT & occ. bias \\
GAP \citep{webster-etal-2018-mind} & 8 908 ex. & Wikipedia & b  & cor. res. & stereotyping \\
KNOWREF & 8 724 sent. & Wikipedia \& other & b & cor. res. & stereotyping \\
BiosBias \citep{dearteaga-2019-bios} & 397 340 bios & CommonCrawl & b & classification & occ. bias \\
GeBioCorpus  & 2 000 sent. & Wikipedia & b & MT & occ. bias \\
StereoSet \citep{nadeem2020stereoset} & 2 022 sent. & human-generated & b & probing LMs & stereotyping \\
CrowS-Pairs & 1508 ex. & human-generated & b & probing LMs & stereotyping \\
\bottomrule
\end{tabular}
}
\caption{List of common probing datasets for gender bias in language.}
\label{tab:dataset}
\vspace*{-0.8cm}
\end{table}


In order to measure gender bias in NLP methods and downstream applications, a number of datsets have been developed. 
We list the well-established datasets in Table \ref{tab:dataset} together with the tasks they can probe and biases they provide a testbed for. Below we discussed three groups of datasets: those based on simple template structures, those based on natural language data, and datasets that have been developed to detect gender bias in language models. 

\subsubsection{Template-Based Datasets}
\label{sec:datasets-template}

A number of studies accounting for gender bias in natural language processing have been conducted on benchmark datasets consisting of template sentences of simple structures such as ``\textit{He/She is a/an [occupation/adjective].}'' where \textit{[person/adjective]} is populated with occupations or positive/negative descriptors \citep{prates2019assessing, cho-etal-2019-measuring,bhaskaran-bhallamudi-2019-good,saunders-byrne-2020-reducing}. 
Similarly, the EEC dataset \citet{kiritchenko2018examining} includes sentence templates such as \textit{[Person] feels [emotional state word].} and \textit{The [person] has two children}. 
The EEC dataset has been widely used in other projects \citep{bhardwaj2020investigating} and has been extended with German sentences by \citet{bartl-etal-2020-unmasking}. 
Another multilingual dataset has been proposed by
\citet{nozza-etal-2021-honest} that create a template-based dataset in 6 languages (English, Italian, French, Portuguese, Romanian, and Spanish) similarly consisting of a subject and a predicate. 


Another strain of work has utilised the structure of Winograd Schemas \citep{levesque-2012-winograd}: WinoBias \citep{zhao-etal-2018-gender}, WinoGender \citep{rudinger-etal-2018-gender}, and WinoMT \citep{stanovsky-etal-2019-evaluating}. Since Winograd Schema Challenge is a coreference resolution task with human-generated sentence templates which requires reasoning with commonsense knowledge, it has been employed to analyse if reasoning of coreference system is dependent on a gender of a pronoun in a sentence and to measure stereotypical and non-stereotypical gender associations for different occupations. 

WinoBias \citep{zhao-etal-2018-gender} contains two types of sentences that require the linking of gendered pronouns to either male or female stereotypical occupations. None of the examples can be disambiguated by the gender of the pronoun, but this cue can potentially distract the model. The WinoBias sentences have been constructed so that, in the absence of stereotypes, there is no objective way to choose between different gender pronouns. 
In parallel, \citet{rudinger-etal-2018-gender} develop a WinoGender dataset \citep{levesque-2012-winograd}. 
As in the WinoBias dataset, each sentence contains three variables: \textit{occupation}, \textit{person} and \textit{pronoun}. For each occupation, Winogender includes two similar sentence templates: one in which \textit{pronoun} is coreferent with \textit{occupation}, and one coreferent with \textit{person}. 
Notably, WinoGender sentences unlike WinoBias also include gender-neutral pronouns. Finally, sentences in WinoGender are not resolvable from syntax alone, unlike in the WinoBias dataset, which might enable better isolation of the effect of gender bias. Both of these datasets have been employed in a number of analysis on gender bias in coreference resolution \citep{jin-etal-2021-transferability, de-vassimon-manela-etal-2021-stereotype, tan2019assessing,vig2020causal}.  

Building on WinoGender and WinoBias, \citet{stanovsky-etal-2019-evaluating} curate WinoMT, a probing dataset for machine translation, with sentences with stereotypical and non-stereotypical gender-role assignments. 
WinoMT has become widely applied as a challenge dataset for gender bias detection in MT systems \citep{stafanovics-etal-2020-mitigating, basta-jussa-2020-mitigating, saunders-byrne-2020-reducing,renduchintala-etal-2021-gender} with \citet{saunders-etal-2020-neural} developing a version of the WinoMT dataset with binary templates filled with singuar \textit{they} pronoun.
Similarly, the Occupations Test dataset \citep{escude-font-costa-jussa-2019-equalizing} contains template sentences to test MT systems on.
Ultimately, both Occupations Test and WinoMT test if the grammatical gender of the translation is aligned with the gender of the pronoun in the original sentence which limits the aspects of gender bias they can probe for. 

\subsubsection{Natural Language Based Datasets}

Probing datasets utilise also available natural language resources and extend them with annotations to tune it for the gender bias detection task. 
Importantly, these datasets can be applied to analyse gender bias in natural language and in algorithms, and are not limited by artificial structures of the template-based approaches to collecting data. 

A number of popular datasets rely on data collected from Wikipedia. For instance, GAP \citep{webster-etal-2018-mind} is a  human-labeled corpus derived from Wikipedia including sentences relevant for coreference resolution task. Unlike WinoGender and WinoBias, GAP focuses on relations where the antecedent is a named entity instead of pronouns \citep{webster-etal-2018-mind} and thus, can be used to unravel biases towards entities. 
Similarly, to analyse gender bias in coreference resolution, \citet{emami-etal-2019-knowref} develop the KNOWREF dataset, which is scraped from Wikipedia together with OpenSubtitles, and Reddit comments. Then, after initial filtering they infer the genders of antecedents 
based on their first names and ask human annotators to predict which antecedent was the correct coreferent of the pronoun. 
Due a relatively large size of these datasets, both GAP and KNOWREF can be used as an alternative to sentence template based datasets.

Another line of work is analysing gender bias in biographies. \cite{dearteaga-2019-bios} develop the BiosBias dataset, which consists of biographies with labelled occupations and gender identified within Common Crawl. 
The dataset has been created for the task of correctly classifying the subject’s occupation from their biography assuming that there are differences between mens' and womens' online biographies other than gender indicators \citet{dearteaga-2019-bios}. 
Further, GeBioCorpus \citep{costajussa2019gebiotoolkit} present a dataset with biography and gender information from Wikipedia which has been widely used to analyse gender bias in MT (for English, Spanish, and Catalan) \citep{vanmassenhove-etal-2018-getting, escude-font-costa-jussa-2019-equalizing, basta-jussa-2020-mitigating}. 

Datasets employ also other online data sources.
For instance, RtGender \citep{voigt-etal-2018-rtgender} is a dataset of online communication to enable research in communication directed to people of a specific gender. Studies on detecting misogynist or toxic
language on social media released Twitter-based datasets  \citep{anzovino2018, hewitt2016}. 
\citet{bentivogli-etal-2020-gender} develop MuST-SHE,
a multilingual benchmark based on TED data for gender bias detection in machine and speech translation.
Recently, \citet{marjanovic2021quantifying} create a dataset with Reddit comments to study gender biases that appear in online political discussion.

\subsubsection{Probing Language Models}

A significant, though relatively recent and thus undiscovered, research direction has concentrated on analysing gender bias in language models. To this end, specific datasets have been curated.
In particular, \citet{nadeem2020stereoset} present StereoSet, which is a dataset to measure stereotypical biases in gender, among other domains. It consists of triplets of sentences with each instance corresponding to a stereotypical, anti-stereotypical or a meaningless association. This dataset enables ranking language models based on probabilities they assign to each of these triplets.  
In parallel, \citet{nangia-etal-2020-crows} introduce CrowS-Pairs, a crowdsourced, template-based challenge set for measuring social biases, including gender bias, that are present in current language models. In CrowS-Pairs, each example consists of a pair of sentences, a stereotypical and anti-stereotypical. 
Both of these datasets are a significant starting point for creating a benchmark for evaluating gender bias in language models. 
Notably, \citet{stanczak2021quantifying} propose a method for generating multilingual
datasets for analysing gender bias towards named entities in LMs.


\subsection{Summary}

Above we have discussed popular datasets employed for analysing gender bias. We note that datasets based on simple template structures allow for a controlled experiment environment. However, we warn that the limitations they impose might include artificial biases, and the results of models tested on them may not map to a more natural environment. Since the above datasets provide means of conducting diagnostic tests for gender bias, they have a high positive and low negative predictive value for the presence of gender bias \citep{rudinger-etal-2018-gender}. Therefore, using these datasets, it is only possible to demonstrate the presence of gender bias in a system but not to prove its absence. Although datasets based on natural language obviate the downsides of the benchmark datasets with simple patterns, they often concentrate on data from one domain, e.g., social media, Wikipedia, or news. Therefore, the results might not generalise well to other domains and should be treated with caution. We note that natural language data might encode gender bias itself so that it is impossible to isolate bias from the data and the tested model. For instance, \citet{chaloner-maldonado-2019-measuring} find evidence of bias in word embeddings trained on the GAP dataset when testing on a standard bias benchmark. They assume that this is due to gender bias on Wikipedia, GAP's underlying data.

However, irrespectively if based on natural language or sentence templates, most of these lexica and datasets are only available for English. Only datasets to analayse gender bias in machine translation, due to the nature of the task, are available in other languages. However, they often consider high-resource languages such as Spanish or German. Similarly, most of these datasets restrict themselves to the binary view on gender presenting a major gap in the research. Thus, we encourage data collection for gender inclusive task-specific datasets. Further, many of the popular publications have focused solely on occupational biases without accounting for a nuanced nature of gender bias. Finally, despite a number of datasets curated specifically to assess for gender bias, only a few can be considered as benchmarks for a targeted downstream task and they come predominantly from the machine translation and coreference resolution domain. Therefore, we strongly encourage further research along the lines of establishing evaluation benchmarks for the underlying models such as \citet{nadeem2020stereoset, nangia-etal-2020-crows}.

\section{Defining bias}
\label{sec:definition}

In the following, we list the common formal definitions of bias that are utilised to quantify the social concepts presented in Section \ref{sec:bias} and divide them into definitions used for detecting gender bias in language (\S \ref{sec:def-nl}), either natural or generated, and in NLP methods (\S \ref{sec:def-algo}). 

\subsection{Measuring Gender Bias in Language}
\label{sec:def-nl}

Gender bias manifests itself in texts in many ways and can be identified using both linguistic and extra-linguistic cues \citep{marjanovic2021quantifying}. Already structure of the data, e.g., the distribution of genders mentioned in the text, can be a bias indicator and the differences in these distributions can be used as a measure for bias. 
However, in the following, we focus on more complex textual biases, \textit{i.e.}, lexical biases, and discuss measures for quantifying differences in portrayals of genders, and their stereotypical depictions.

\subsubsection{Differences in Gender Descriptions}

Differences in depictions of men and women have been prolificly quantified using point-wise mutual information (PMI) \citep{rudinger-etal-2017-social,hoyle-etal-2019-unsupervised,stanczak2021quantifying}. In particular, PMI investigates the co-occurrence of words with a particular gender. In PMI descriptors (such as adjectives or verbs) linked to a gendered entity are counted and the probability of their co-occurrence to a gender across entity is calculated. More formally, PMI is defined as:

\begin{equation}
PMI(gender, \textbf{word}) = ln \bigg( \frac{P(gender,\textbf{word})}{P(gender)P(\textbf{word})}\bigg)
\label{eq:PMI}
\end{equation}

In general, words with high PMI values for one gender are suggested to have a high gender bias. However, \citet{rudinger-etal-2017-social} note that bias at the level of word co-occurrences is likely to lead to overgeneralisation when applied to a heterogenous dataset. Notably, PMI can also be used to measure differences in word choice for genders beyond the binary 
\citep{stanczak2021quantifying}.

Further, \citet{hoyle-etal-2019-unsupervised} extend the PMI approach and propose an unsupervised model that jointly represents descriptors with their sentiment to investigate gender bias in words used to describe men and women together with word's sentiment.


\subsubsection{Stereotypical and Occupational Bias}

Occupational gender segregation and stereotyping is a major problem in the labor market often caused by gender roles and stereotypes present in society and as such has been in focus in a numerous research \citep{lu2019gender}. To this end, \citet{qian-2019-gender} calculate an overall stereotype score of a text as the sum of stereotype scores of all the by definition gender-neutral words 
with gendered words in the text, divided by the total count of words calculated. 
Then, \citet{qian-2019-gender} define the gender stereotype score of a word:
\begin{equation*}
    bias(\textbf{word}) = \bigg\lvert \log \frac{c(\textbf{word}, m)}{c(\textbf{word}, f)} \bigg\rvert
\end{equation*}
where $f$ is a set of female words (e.g., she, girl, and woman), and $c(\textbf{word}, g)$ is the number of times a gender-neutral $\textbf{word}$ co-occurs with gendered words. A word is used in a neutral way, if the stereotype score is 0, which means it occurs equally frequently with male words and females word in the text. \citet{qian-2019-gender} assess occupation stereotypes score in a text as the average stereotype score of a list of gender-neutral occupations in the text. 
This definitions of stereotypical and occupational bias have been employed in subsequent research \citep{bordia-bowman-2019-identifying, qian-etal-2019-reducing}. 

\subsection{Measuring Gender Bias in Methods}
\label{sec:def-algo}
With the prevalence of NLP systems and their increasing application areas, researchers have developed measures to probe for gender biases encoded in these methods. In the following, we discuss different definitions used for bias detection in NLP methods.

\subsubsection{Bias influencing Performance}

For downstream tasks where there exists a gold gender, researchers have utilised performance-based measures to quantify bias. In particular, these measures are relevant for applications such as machine translation and coreference resolution where the objective involves the correct handling of gendered (pro-)nouns. 

Then, the amount of bias encoded in NLP systems can be quantified using: accuracy (percentage of observations with the correctly gendered entity) \citep{saunders-byrne-2020-reducing}; difference in accuracy between the set of sentences with anti-stereotypical and stereotypical sentences; $F_1$ score and difference in $F_1$ score between the stereotypical and anti-stereotypical gender role assignments \citep{zhao-etal-2018-gender,webster-etal-2018-mind,de-vassimon-manela-etal-2021-stereotype}; log-loss of the probability estimates \cite{webster-etal-2019-gendered}; false positive rates \citep{kennedy-etal-2020-contextualizing,jin-etal-2021-transferability}; ratio of observations with masculine and feminine predictions; gender differences in distributions of and within occupations \citet{kirk2021bias}.


Depending on the downstream task, task-specific performance measures are used to evaluate gender bias. For instance, to assess gender bias in dependency parsing, the labeled attachment score that measures the percentage of tokens that have a correct assignment 
and the correct dependency relation has been applied \citep{garimella-etal-2019-womens}. Next, BLEU is used in machine translation to assess the quality of the translated text \citep{saunders-byrne-2020-reducing}. If the MT system is gender biased, the system produces an incorrect gender predicition even when no ambiguity exists \citep{costa-jussa-de-jorge-2020-fine}. Thus, the lower the bias, the better the translation quality in terms of BLEU score and accuracy \citep{escude-font-costa-jussa-2019-equalizing, stanovsky-etal-2019-evaluating,basta-jussa-2020-mitigating}. However, \citet{bentivogli-etal-2020-gender} point out that previously obtained BLEU gains \citep{vanmassenhove-etal-2018-getting, moryossef-etal-2019-filling} cannot be ascribed with certainty to a better control of gender features and following previous research \citep{elaraby2018gender, vanmassenhove-etal-2018-getting} underlie the importance of applying gender-swapping in BLEU-based evaluations focused on gender translation.



\subsubsection{Stereotypical Bias}

Another stream of research attempts to quantify gender bias in terms of stereotypical associations that a method conveys. For instance, \citet{zhao-etal-2018-gender} consider a system gender biased if it links pronouns to occupations more accurately for the stereotypical pronoun, rather than the anti-stereotypical one. 
Next, in order to assess stereotypical associations encoded in NLP methods, \citet{kurita-etal-2019-measuring} suggest to measure how much more a model prefers the male association with a certain attribute, e.g., a programmer, compared to the female gender. To this end, \citet{kurita-etal-2019-measuring} propose to create template sentences, similar to the ones discussed in \S \ref{sec:datasets-template}, 
and calculate a log probability bias score for BERT predictions when filling in a template with the gendered words and the target word. This measure has been widely applied in numerous research \citep{bartl-etal-2020-unmasking, vig2020causal}.
Building up on this approach, \citet{munro-morrison-2020-detecting} calculate the ratio of the actual probabilities instead of log probabilities, claiming that ratios allow for more transparent comparisons. 


For datasets where each instance contains at least two versions of the same template sentence, e.g., male and female, the paired t-test has been used to measure if the mean predicted class probabilities are different across genders \citep{kiritchenko2018examining,bhaskaran-bhallamudi-2019-good}. Similarly, \citet{nangia-etal-2020-crows} propose a metric that calculates the percentage of examples for which the language model is in favor of the more stereotyping sentence. To measure this, \citet{nangia-etal-2020-crows} first break each sentence in an example into two parts: the modified tokens that appear only in one of the sentences and the unmodified part that is shared. Then, using pseudo-log-likelihood masked language model scoring \citep{salazar-etal-2020-masked}, they estimate the probability of the unmodified tokens conditioned on the ones.

Due to their simplicity and interpretability the above measures have been widely adopted to measure gender bias. However, these methods cover only stereotypical bias neglecting many other ways in which gender bias can be expressed.  

\subsubsection{Causal Bias}

Causal testing presents another way of measuring gender bias in NLP systems. Then, gender bias is defined as the disparity in the output when model is feeded with different genders \citep{qian-etal-2019-reducing}.
\citet{lu2019gender} define bias as the expected difference in scores assigned to expected absolute bias across different genders. Later, \citet{qian-etal-2019-reducing} limit the above bias evaluation to a set of gender-neutral occupations and measure how the probabilities of occupation words depend on the gendered word and in reverse, how the probabilities of gendered wordsdepend on the occupation words. Similarly, \citet{emami-etal-2019-knowref} propose consistency as a bias metric, where they duplicate the dataset by switching the candidate antecedents each time they appear in a sentence. If a coreference model relies on knowledge and contextual understanding, its prediction should differ between the two versions. 
\citet{emami-etal-2019-knowref} define the consistency score as the percentage of predictions that change from the original instances to the switched instances.

Causal testing in gender bias detection has been used to define bias in terms of stereotypical bias, rather than approaching other possible harms, which sets a possible ground for future work.



\subsubsection{Male Default}

Gender bias can be defined as the deviation of the distribution of gender pronouns in an output of an NLP system from a gender distribution of demographics of an occupation \citep{prates2019assessing}. These differences occur more often in a presence of the male default phenomenon (\S \ref{sec:bias}). Especially in machine translation systems, male defaults lead to overestimating the distribution of male instances over female ones. 

To account for male default in MT, \citet{cho-etal-2019-measuring} propose a translation gender bias index (TGBI) and apply it to Korean-English translations. Let $p^{f}_{i}$ be the portion of a sentence translated to a female pronous, $p^{m}_{i}$ as male and $p^{n}_{i}$ as gender-neutral pronouns in any set of sentences $S_{i} \in S$.
\begin{equation*}
    TGBI = \frac{1}{n}\sum_{i=1}^{n}\sqrt{p^{f}_{i} p^{m}_{i} + p^{n}_{i}}
\end{equation*}
where 
$p^{f}_{i} + p^{m}_{i} + p^{n}_{i} = 1$ and $p^{f}_{i}, p^{m}_{i}, p^{n}_{i} \in [0, 1]$ for each $i$. TGBI is equal to 1 in optimum when all the predictions incorporate gender-neutral terms. \citet{cho-etal-2019-measuring} expect TGBI to be a representative measure for inter-system comparison, especially if the gap between the systems is noticeable. Recently, \citet{ramesh-etal-2021-evaluating} extend TGBI to Hindi. In general, this is a suitable method for applications where male default is the predominant risk. 



\subsubsection{Bias in Word Embeddings}

In recent years, a myriad of publications have approached quantifying bias in word embeddings. In the following, we present the according to our judgement most influential research in this field. 

\paragraph{Projection-Based Measures} 
In the initial work on gender bias in word embeddings, \citet{bolukbasi2016man} distinguish between two types of bias, direct and indirect.
Following \citet{bolukbasi2016man} direct bias of a word embedding $\overrightarrow{w}$ can be quantified as:
\begin{equation*}
    DirectBias_c = \frac{1}{\mid N \mid} = \sum_{w \in N}\mid cos(\overrightarrow{w}, g) \mid^
c\end{equation*}
where $N$ is a set of gender neutral words, $g$ is the gender direction and $c$ is a parameter determining how strict bias is defined. The direct bias manifests itself in relative similarities between gendered and gender-neutral words. However, since gender bias could also affect the relative geometry between gender neutral words themselves, \citet{bolukbasi2016man} introduce notion of indirect gender bias which manifests as associations between
gender neutral words that are arising from gender.
In particular, if word such as \textit{businessman} and \textit{genius} are closer to \textit{football}, a word with an embedding closer in the gender subspace to a man, it can indicate indirect gender bias. However, \citet{gonen-goldberg-2019-lipstick} argue that the indirect bias has been disregarded to some extent and complain that mitigation methods are not provided.   

Another researched distance-based metric to measure gender bias in word embeddings uses the relative norm distance between two groups \citep{Garg_2018}:
\begin{equation*}
d = \sum_{v_m \in M} \lVert v_m - v_1 \rVert_2 - \lVert v_m - v_2 \rVert_2
\end{equation*}
where $M$ is the set of neutral word vectors and $v_i$ is the average vector for group $i$. The more positive (negative) that the relative norm distance is, the more associated the neutral words are towards group two (one). Thus, the above metric captures the relative distance (\textit{i.e.}, relative strength of association) between the group words and the neutral word list of interest. Similarly, \citet{friedman-etal-2019-relating} compute bias as the average axis projection of a neutral word set onto the male-female axis and evaluate it for any region's word embedding computing its correlation to gender gaps. 

Since the above definitions are straightforward and geometrically grounded, they have been often employed to quantify gender bias in word embeddings. 
However, bias is much more profound and systematic than the projection of words \citep{gonen-goldberg-2019-lipstick}. 

\paragraph{Word Embedding Association Test (WEAT)} 
The WEAT 
has been developed as a benchmark for testing gender bias in word embeddings via semantic similarities.
In particular, the WEAT compares set of target concepts (e.g., male and female words) denoted as $X$ and $Y$ (each of equal size $N$), with a set of attributes to measure bias over social attributes and roles (e.g., career/family words) denoted as $A$ and $B$.
The resulting test statistics is defined as a permutation test over $X$ and $Y$:
\begin{equation*}
    S(X, Y, A, B) = [mean_{x \in X}sim(x, A, B) -  mean_{y \in Y}sim(y, A, B)]
\end{equation*}
where $sim$ is the cosine similarity. The resulting effect size is then the measure of association:
\begin{equation*}
    d = \frac{S(X, Y, A, B)}{std_{t \in X \cup Y}s(t, A, B)}
\end{equation*}

The null hypothesis suggests there is no difference between $X$ and $Y$ in terms of their relative similarity to $A$ and $B$. In \citet{Caliskan_2017}, the null hypothesis is tested through a permutation test, \textit{i.e.}, the probability that there is no difference between $X$ and $Y$ (in relation to $A$ and $B$) and therefore, that the word category is not biased. 
However, we note that results obtained with WEAT should be treated with a grain of salt since \citet{ethayarajh-etal-2019-understanding} prove that WEAT systematically overestimates bias.


\paragraph{Sentence Embedding Association Test (SEAT)}
Based on the WEAT, \citet{may-etal-2019-measuring} develop an analogous method,  
SEAT, that compares sets of sentences, rather than words.
In particular, \citet{may-etal-2019-measuring} apply WEAT to the sentence representation. Thus, WEAT can be seen as a special case of SEAT in which the sentence is a single word. To extend a word-level test to sentence contexts, \citet{may-etal-2019-measuring} slot each word into each of several semantically bleached sentence templates.  


\paragraph{Bias Amplification}
Previous research has shown that NLP models are able not only to perpetuate biases extant in language, but also to amplify them \citep{zhao-etal-2017-men}. In particular, \citet{zhao-etal-2017-men} interpret gender bias as correlations that are potentially amplified by the model and define gender bias towards a $man$ for each word as:
\begin{equation}
    b(word, man) = \frac{c(word, man)}{c(word, man) + c(word, woman)}
\end{equation}
where $c(word, man)$ is the number of occurrences of a word and male gender in a corpus.
If $b(word, man) > 1/\lvert G \rvert$ ($G = \{man, woman\}$ under gender binarity assumption), then a word is positively correlated with gender and may exhibit bias. To evaluate the degree of bias amplification, \citet{zhao-etal-2017-men} propose to compare bias scores on the training set, $b^{*}(word, man)$, with bias scores on an unlabeled evaluation set.
We note that this method is applicable solely to individual words and would require an extension to be used as a general evaluation metric. 




\subsubsection{Qualitative Assessment}

Alongside the above discussed quantitative gender bias measures, some research includes qualitative measures to analyse the extent of gender bias. For instance, \citet{moryossef-etal-2019-filling} conduct a syntactic analysis of generated translations examining inflection statistics for sentence templates from the dataset. \citet{escude-font-costa-jussa-2019-equalizing} introduce clustering as a measure of gender bias. Then, the higher the clustering accuracy for stereotypically-gendered words, the more bias the word embeddings trained on the dataset have. We find this line of work particularly interesting as it encourages better model understanding and interpretability.

\subsection{Summary}


Gender bias can be expressed in language in many nuanced ways which poses stating a comprehensive definition as one of the main challenges in this research field. In this section, we have examined different gender bias definitions. We find that they vary dramatically across and within algorithms and tasks, which supports findings made by \citet{blodgett2020language} that analyse bias definitions in general. Bias is often described only implicitly without any formal definition. Even when a paper states a formal definition, it essentially covers only one type of bias which oversimplifies the task and thus, makes it impossible to detect all harmful signals in language. In particular, we discuss a number of methods to quantify bias in word embeddings which are utilised in many downstream tasks. However, most of them consider only one way of defining bias and do not engage enough parallel research to combine these methods. We here support \citep{silva-etal-2021-towards}'s claim that solely using one bias metric or test is not enough -- diversifying metrics to ensure robustness of the evaluations is thus important. 
Additionally, we strongly encourage developing standard evaluation measures and tests to enhance comparability.    

Another limitation we see is that defining bias in terms of decreasing performance, however straight-forward, carries a risk of capturing bias only as long as it influences the performance. This way bias detection is only a means of enhancing model's performance instead of being a goal on its own which can raise ethical considerations. Moreover, some of the performance measures have been previously criticised as evaluation benchmarks for tasks they address. For instance, it is widely acknowledged in machine translation that BLEU score is a coarse and indirect indicator of a machine translation system's performance \citep{callison-burch-etal-2006-evaluating}.  

Finally, similarly to our observations regarding datasets, most of the measures developed for quantifying gender bias are created and calculated only for binary genders. Even if a specific metric allows for analysing non-binary genders, it usually remains unmentioned.



\section{Detecting Gender Bias}
\label{sec:detection}

Armed with datasets (\S\ref{sec:resources}) suitable for gender bias analysis and formal gender bias definitions (\S\ref{sec:definition}), we focus herein on research on detecting and analysing the nature of gender bias in natural language, NLP algorithms, and downstream tasks. We discuss its challenges, and influential lines of work. 


\subsection{Detecting Gender Bias in Natural Language}

Natural language is known to exhibit societal biases. Gender bias, in particular, has been studied in a broad spectrum of texts such as portrayals of characters in movies, books, news and media. 

\citet{ramakrishna-etal-2017-linguistic, ramakrishna-etal-2015-quantitative, Choueiti2014GenderBW} examined gender differences in portrayal of characters in movies and consistently show that female characters appear to be more positive in language use with fewer references to death and fewer swear words compared to male characters. However, \citet{sap-etal-2017-connotation} find that, high-agency women frames are rare in modern films. \citet{rashkin-etal-2018-event2mind} use commonsense inference tasks on movie scripts' corpus to unveil presence of gender bias finding that women's looks and sexuality are highlighted, while men's actions are motivated by violence, with strong negative reactions. Moreover, \citet{bamman-smith-2014-unsupervised} employ a probabilistic latent-variable model to extract event classes from biographies and find that characterisation bias on Wikipedia with biographies of women containing significantly more emphasis on events of marriage and divorce than biographies of men. 
\citet{field-tsvetkov-2019-entity} 
show that although powerful women are frequently portrayed in the media, they are typically described as less powerful than their actual role in society. However, \citet{asr2021} report that there, in fact, is a gender gap in coverage of women in Canadian news outlets.  
Further, \citet{hoyle-etal-2019-unsupervised} use an unsupervised model to find that 
differences between descriptions of males and females in literature align with common gender stereotypes: Positive adjectives used to describe women are more often related to their bodies than adjectives used to describe men.

However, \citet{Garg_2018} show that gender bias has decreased in the last 100 years and that the women’s movement in the 1960s and 1970s had a significant effect on women's portrayals in literature and culture. To this end, \citet{Garg_2018} use word embeddings as a tool to observe the development of adjectives associated with men and women. This is possible since word embeddings learn harmful associations and stereotypes from the underlying data and thus, may serve as a means to extract implicit gender associations from a corpus to detect gender associations present in society \citep{bolukbasi2016man}.
Similarly, \citet{wevers-2019-using} show that word embeddings can be used to investigate shifts in language related to gender,
while \citet{friedman-etal-2019-relating} prove that word embeddings are able to characterise and predict statistical gender gaps in education politics, economics and health across cultures.

A number of research has investigated differences in language directed towards men and women. For instance, \citet{Tsou2014ACO} find that comments on TED talks are more likely to be about the presenter than the content if the presenter is a woman. \citet{fu2016tiebreaker} analyse questions directed at male and female tennis players, finding that questions to men are rather about the game while questions directed at women are often about their appearance and relationships. Further, \citet{voigt-etal-2018-rtgender} corroborate the former findings, such as remarks on appearance being more often targeted towards women, responses to women being more emotive (non-neutral sentiment) and of higher sentiment in general which can be ascribed to benevolent sexism. 

While the above research unveils some of the ways gender bias is manifested in natural language, it gives only a limited view since most of this research has concentrated on binary gender identities and was mostly conducted in English. We note that there exist real-life applications with societal implications to algorithms detecting gender bias in natural language such as warning systems classifying texts as biased to notify readers. 




\subsection{Detecting Gender Bias in Methods}


Biased datasets used in the training process are the primary source of gender bias in NLP methods \citep{zhao-etal-2017-men}. 
\citet{zhao-etal-2019-gender, tan2019assessing} examine datasets that were used as training corpora for the popular NLP methods and find that the occurrence of male pronouns is consistently higher across all datasets and evidence of stereotypical associations. These gender imbalances lead to gender bias in the NLP systems, such as coreference resolution \citet{zhao-etal-2018-gender}. 
It has been shown that the level of bias encoded in a model differs depending on the training data. For instance,  \citet{chaloner-maldonado-2019-measuring} study differences in bias in a number of word embeddings trained on corpora from four domains showing the lowest bias in word embeddings trained on a biomedical corpus and the highest bias when trained on news data (higher than social media and Wikipedia-based corpus). Surprisingly, \citet{lauscher-glavas-2019-consistently}'s findings confirm that gender bias seems to be less pronounced in embeddings trained on social media texts.

A common phenomenon leading to gender bias is a generic masculine pronoun which arises when the masculine form is taken as the generic form to designate all persons of any gender. This is especially the case in the gendered languages \citep{carl-etal-2004-controlling}. Generic masculine poses a challenge in text interpretation since it is unclear if a given person denotation refers to a particular person or a generic form to describe all people in a specific group. For instance, in a sentence \textit{``A researcher must always test his model for biases.''}, it is ambiguous if a particular researcher is considered or researchers in general. In particular, \citet{hitti-etal-2019-proposed} analyse data from Project Gutenberg and IMDB to identify such gender generalisations and detect that even 5\% of each corpus is affected. 

Due to simple interpretation and ability to capture gender stereotypes occupation words have become a common domain for gender bias detection \citep{Garg_2018}. \citet{bolukbasi2016man} project the occupation words onto the \textit{she-he} axis and find that the projections are strongly correlated with the stereotypicality estimates of these words, suggesting that the geometric bias of word embeddings is aligned with crowd judgment of gender stereotypes. \citet{sahlgren-olsson-2019-gender} show that male names are on average more similar to stereotypically male occupations with an according observation applying to female names. \citet{rudinger-etal-2018-gender} demonstrate how occupation-specific bias is correlated with employment statistics and often so magnified.

Although the majority of the research has focused on analysing gender bias in methods developed on English corpora, there have been some advances in extending this line of work to other languages. Developing language-specific methods to assess language model's limitations is crucial to prevent bias propagation to downstream tasks in the analysed language \citep{sun-etal-2019-mitigating, bartl-etal-2020-unmasking}. Findings made for English do not automatically extend to other languages, especially if those exhibit morphological gender agreement \citep{nozza-etal-2021-honest}. 
In particular, gender bias in word embeddings of languages with grammatical gender can be expressed in different ways, such as in a discrepancy in semantics between the masculine and feminine forms of the same noun in word embeddings. For example, it has been shown that when aligning Spanish to English word embeddings, the word ``abogado'' (male lawyer) is closer to ``lawyer'' than ``abogada'' (female lawyer) \citep{zhou-etal-2019-examining}. 
Interestingly, \citet{lauscher-glavas-2019-consistently} find that the level of bias in cross-lingual embedding spaces can roughly be predicted from the bias of the corresponding monolingual embedding spaces. 

Model architecture is analysed as one of the influencing factors for bias in algorithms. For instance, \citet{lauscher-glavas-2019-consistently} hypothesise that the bias effects reflected in the distributional space depend on the preprocessing steps of the embedding model. Additionally, discovering bias in transformer models has proven to be more nuanced than bias-discovery in word embedding models \citep{kurita-etal-2019-measuring, may-etal-2019-measuring}. \citet{nadeem2020stereoset} hyphotesise that an ideal language model should not only be able to perform the task of language modeling, but also cannot exhibit stereotypical bias -- it should avoid ranking stereotypical contexts higher than anti-stereotypical contexts. Recent research has aimed to rank language models in terms of bias they perpetuate \citep{nangia-etal-2020-crows,silva-etal-2021-towards}. However, these studies present partially contradictory results presenting a need for more exhaustive testing. 
The influence of the model's size on the encoded (gender) bias has been examined. For instance, \citet{silva-etal-2021-towards} find that distilled models almost always exhibit statistically significant bias and that the bias effect sizes are often much stronger than in the original models. 
\citet{vig2020causal} show that gender bias increases with the size of a model. Recently, \citet{bender2021dangers} confirm this claim warning from potential risks associated with large language models. However, in a study of gender bias in cross-lingual language models \citet{stanczak2021quantifying} do not find significant results to support this claim.  


 

It is difficult to understand the nature of biases encoded in large language models due to their complexity. However, applying interpretability methods can shed light on the models and biases preserved. For instance, \citet{vig-2019-multiscale} use visualisations to reveal attention patterns generated by GPT-2 in the task of conditional language generation and show that the model's coreference resolution might be biased. 
\citet{vig2020causal} probe neural models to analyse the role of individual neurons and attention heads in mediating gender bias and find out that the source of gender bias is concentrated in a small part of the model. Moreover, \citet{bhardwaj2020investigating} identify gender informative features (and discard them from the model as a mitigation technique).

Until now research has aimed to detect gender bias in a strictly binary setting. We want to highlight the importance of a gender-inclusive research and discuss below publications that have step up to this task. \citet{hicks2015} collect a data set and develop visualisation tools that show relative frequency and co-occurrence networks for American English trans words on Twitter. \citet{manzini-etal-2019-black} extend the method presented in \citet{bolukbasi2016man} and use their approach to find non-binary gender bias by aggregating n-tuples instead of gender pairs. 
\citet{saunders-etal-2020-neural} explore applying tagging to indicate gender-neutral referents in coreference sentences with a gender-neutral pronoun.
Recently, \citet{vig2020causal} test the probability of a model to generate the pronoun \textit{they} for a number of templates. The probability of the pronoun \textit{they} is relatively low, however constant across probed professions.


\subsection{Detecting Gender Bias in Downstream Tasks}
\label{sec:detect-tasks}

Bias in the above methods influences many downstream tasks for which these methods are used, which presents a risk of propagating and amplifying gender bias \citep{zhao-etal-2017-men, zhao-etal-2018-gender}. Thus, in the following, we analyse literature on gender bias in downstream applications. 

\paragraph{Machine Translation}

Popular online machine learning services, such as Google Translate or Microsoft Translator, were shown to exhibit biases and to default to the masculine pronoun \citep{escude-font-costa-jussa-2019-equalizing}. 
NLP models may learn associations of gender-specified pronouns (for a gendered language) and gender-neutral ones for lexicon pairs that frequently collocate in the corpora \citep{cho-etal-2019-measuring}. This kind of phenomenon threatens the fairness of a translation system since it lacks generality and inserts social bias to the inference. Moreover, the output is not fully correct (considering gender-neutrality) and poses ethical considerations.

When translating from a language without grammatical gender to a gendered one, the required clue about the noun's gender is missing which poses a challenge for MT systems. 
\citet{saunders-etal-2020-neural} find that existing approaches tend to overgeneralise and incorrectly use the same inflection for every entity in the sentence.
However, gender is incorrectly predicted not only in the absence of the gender information. MT methods produce stereotyped translations even when gender information is present in the sentence.
\citet{schiebinger2014gender} argue that scientific research fails to take this issue into account. 
Recently, \citet{prates2019assessing} show that Google Translate still exhibits a strong tendency towards male defaults, in particular for fields typically associated with unbalanced gender distribution or stereotypes such as STEM (Science, Technology, Engineering, and Mathematics) jobs. \citet{prates2019assessing} hypothesise that gender neutrality in language and communication leads to improved gender equality. Thus, translations should aim gender-neutrality, instead of defaulting to male or female variants.

\paragraph{Coreference Resolution}

Various aspects of gender are embedded in coreference inferences, both because gender can show up explicitly (e.g., pronouns in English, morphology in Arabic) and because societal expectations and stereotypes around gender roles may be explicitly or implicitly assumed by speakers or listeners \citep{cao-daume-iii-2020-toward}. 
Although existing corpora have promoted research into coreference resolution, they suffer from gender bias \citep{zhao-etal-2018-gender}.


\citet{webster-etal-2018-mind} find that existing resolvers do not perform well and are biased to favour better resolution of masculine pronouns. \citet{rudinger-etal-2018-gender} show how overall, male pronouns are more likely to be resolved as occupation than female or neutral pronouns across all systems. Moreover, \citet{zhao-etal-2018-gender} demonstrate that neural coreference systems all link gendered pronouns to stereotypical entities with higher accuracy than anti-stereotypical entities. \citet{zhao-etal-2018-gender} warn that bias encoded in word embeddings leads to sexism in coreference resolution.
Further, \citet{bao-qiao-2019-transfer} show significant gender bias when using popular NLP methods for coreference resolution on both sentence and word level, indicating that women are associated with family while men are associated with career. 

\paragraph{Language Generation}  

\citet{henderson2017ethical} suggest that, due to their subjective nature and goal of mimicking human behaviour, data-driven dialogue models are prone to implicitly encode underlying biases in human dialogue, similar to related studies on biased lexical semantics derived from large corpora \citep{Caliskan_2017, bolukbasi2016man}. \citet{cercas-curry-rieser-2018-metoo} estimate that as many as 4\% of conversations with chatbased systems are sexually charged.  
Further, \citet{bartl-etal-2020-unmasking} find that the monolingual BERT reflects the real-world bias of the male- and female-typical profession groups through stereotypical associations. Stories generated by GPT-3 differ based on a perceived gender of the character in a prompt with female characters being more often associated with family, emotions and appearance, even in spite of a presence of power verbs in a prompt \citep{lucy-bamman-2021-gender}.

\paragraph{Sentiment Analysis} 

\citet{kiritchenko2018examining} test 219 automatic sentiment analysis systems that participated in SemEval-2018 Task 1 \textit{Affect in Tweets} \citep{mohammad-etal-2018-semeval}. In particular, \citet{kiritchenko2018examining} examine a hypothesis that a system should equally rate the intensity of the emotion expressed by two sentences that differ only in the gender of a person mentioned and find that the majority of the systems studied show statistically significant bias. In particular, they consistently provide slightly higher sentiment intensity predictions for sentences associated with one gender (gender with more positive sentiment varies based on a task and system used). When predicting anger, joy, or valence, the number of systems with consistently higher sentiment for sentences with female noun phrases is higher than for male noun phrases. 
\citet{bhaskaran-bhallamudi-2019-good} show that analysed sentiment analysis methods exhibit differences in mean predicted class probability between genders, though the directions vary again.  

\subsection{Summary}


As seen above, NLP methods tend to be consistently biased and associate harmful stereotypes with genders.   
Despite this fact, most of the papers that have focused on detecting gender bias in natural language, methods, or downstream tasks, have seen bias detection as a goal in itself or a means of analysing the nature of bias in domains of their interest. Some of this research has been followed up with bias mitigation methods (discussed in \S\ref{sec:mitigation}). However, often enough, findings of this line of research are treated solely as a fact statement and not an action trigger. In particular, despite a number of evidence showing that NLP methods encode gender bias, developers are not required to provide any formal testing prior to releasing new models. Widely acknowledged models that have led in recent years to significant gains on many NLP tasks have not included any study of bias alongside the publication \citep{conneau-etal-2020-unsupervised,devlin-etal-2019-bert,peters-etal-2018-deep,Radford2019LanguageMA}. Since these models were probed for gender bias only after their release, they might have already caused harm in real life applications. We strongly encourage including bias detection into the model development pipeline and see it as a necessary future development.

So far, research has predominantly aimed to detect bias towards male and female gender, ignoring non-binary gender identities. However, it is crucial to design studies on gender bias detection that are gender-inclusive at all stages, from defining gender and bias, dataset choice to selecting bias detection method.  

As discussed in \S\ref{sec:gender}, gender manifests itself in different ways across languages. Hence, it can be expected that it's also the case for gender bias. For instance, languages such as German, Hebrew and Russian use gender inflections that reflect grammatical genders of the nouns. Further, gender bias is grounded in societal and cultural views on gender and thus, its expressions potentially vary across languages. Expanding research to languages beyond English and including data from outside of the Anglosphere would lead to gaining a broader view on gender bias in societies. In particular, analysing cross-lingual data might enable a comparative studies of gender bias.     







\section{Mitigating Gender Bias}
\label{sec:mitigation}

While it is impossible to altogether remove gender bias from language or from NLP algorithms, research on gender bias mitigation is a significant step towards developing fair systems. In specific applications, one might argue that gender biases in algorithms could capture valuable statistics such as a higher probability of a nurse being a female. Nevertheless, given the potential risk of employing machine learning algorithms that amplify gender stereotypes, \citet{bolukbasi2016man} recommend erring on the side of neutrality and using debiased methods. However, following \citet{dignazio-2021-data-feminism}, mitigating gender bias in AI systems is a short-term solution that needs to be combined with higher-level long-term projects in challenging the current social status quo.

The main challenge in debiasing task is to strike a trade-off between maintaining model performance on downstream tasks while reducing the encoded gender bias \citep{zhao-etal-2018-gender,de-vassimon-manela-etal-2021-stereotype}. 
Further, \citet{sun-etal-2019-mitigating, bartl-etal-2020-unmasking} emphasise the need for more typological variety in NLP research as well as for language-specific solutions. Many of the mitigation methods rely on pre-defined words lists that are not scalable in a multilingual setup and are tedious to create. However, recent work on dictionary definitions for debiasing might obviate the need for predefined word lists \citep{kaneko-bollegala-2021-dictionary}.  
While prior work has mainly focused on mitigating gender bias in English, more recently, researchers have started to apply methods initially developed for English to other languages as well. Naturally, a significant chunk of work for multilingual settings has been researched in the context of neural machine translation \citep{vanmassenhove-etal-2018-getting, prates2019assessing}. This stream of research has confirmed that language-specific solutions are required, since gender is expressed in different ways across languages. For instance, transferring a method successful in gender bias mitigation for English to German may be ineffective which emphasises the need for more typological variety in research as well as language-specific solutions \citep{bartl-etal-2020-unmasking}. Therefore, it is crucial to develop (language-specific) debiasing methods, especially for relatively new methods, to assess these limitations. 
Next, \citet{kiritchenko2018examining} observed that different debiasing approaches have varying effects on the analysed word embedding architectures. 
Many of the current debiasing methods are evaluated only on selected downstream tasks without testing them in a broader scope. Hence, additional and potentially costly tests are required before applying these techniques to other, previously un-tested tasks since their effectiveness there is unclear \citep{jin-etal-2021-transferability}. Therefore, we encourage research on debiasing methods in the early modelling stages.


\begin{table}[h]
\centering
\resizebox{\textwidth}{!}{
\begin{tabular}{llll}
\toprule
\multicolumn{4}{c}{\scalebox{1.5}{\textbf{Data Manipulation}}}
\\  \midrule
\multicolumn{1}{c}{\scalebox{1.3}{\textbf{Data Augmentation}}} & \multicolumn{1}{c}{\scalebox{1.3}{\textbf{Gender Tagging}}} & \multicolumn{1}{c}{\scalebox{1.3}{\textbf{Balanced Fine-Tuning}}} & \multicolumn{1}{c}{\scalebox{1.3}{\textbf{Adding Context}}} \\ \midrule
\citet{park-etal-2018-reducing,Madaan2018AnalyzeDA} & \citet{vanmassenhove-etal-2018-getting,moryossef-etal-2019-filling}  & \citet{park-etal-2018-reducing,saunders-byrne-2020-reducing} & \citet{basta-jussa-2020-mitigating}\\
\citet{zhao-etal-2018-gender,hall-maudslay-etal-2019-name} & \citet{habash-etal-2019-automatic,stafanovics-etal-2020-mitigating} & \citet{costa-jussa-de-jorge-2020-fine} & \\
\citet{zmigrod-etal-2019-counterfactual,emami-etal-2019-knowref} & \citet{saunders-etal-2020-neural} & & \\
\citet{zhao-etal-2019-gender, bartl-etal-2020-unmasking} & & & \\
\citet{de-vassimon-manela-etal-2021-stereotype,sen-etal-2021-counterfactually} & & & \\ \midrule
\multicolumn{4}{c}{\scalebox{1.5}{\textbf{Methodological Adjustment}}} \\ \midrule
\multicolumn{1}{c}{\scalebox{1.3}{\textbf{Projection-Based Debiasing}}}  & \multicolumn{1}{c}{\scalebox{1.3}{\textbf{Adversarial Learning}}}  & \multicolumn{1}{c}{\scalebox{1.3}{\textbf{Constraining Output}}}  & \multicolumn{1}{c}{\scalebox{1.3}{\textbf{Other}}} \\ \midrule
\citet{schmidt2015rejecting,bolukbasi2016man} & \citet{li-etal-2018-towards,zhang2018mitigating} & \citet{zhao-etal-2017-men,ma-etal-2020-powertransformer} &
\citet{zhao-etal-2018-learning,qian-etal-2019-reducing} \\
\citet{bordia-bowman-2019-identifying,park-etal-2018-reducing} & & & \citet{kaneko-bollegala-2019-gender,jin-etal-2021-transferability}\\
\citet{sahlgren-olsson-2019-gender,ethayarajh-etal-2019-understanding} & & & \\
\citet{karve-etal-2019-conceptor,sedoc-ungar-2019-role} & & & \\
\citet{prost-etal-2019-debiasing,liang-etal-2020-monolingual} & & & \\
\citet{Dev_Li_Phillips_Srikumar_2020,kaneko-bollegala-2021-debiasing} & & & \\
\bottomrule
\end{tabular}
}
\caption{Classification of gender bias mitigation methods with respective publications.}
\label{tab:mitigation}
\vspace*{-0.5cm}
\end{table}

Different approaches have been developed to mitigate gender bias in NLP. In this paper, we classify each of these methods following the two main categories, similarly to \citet{sun-etal-2019-mitigating} -- debiasing using data manipulation \ref{sec:debias-corpus} and by adjusting algorithms \ref{sec:debias-algo} -- while extending the scope of our analysis with recent publications and incorporating word embeddings mitigation methods into the methodoligical adjustment category. We summarise the identified lines of gender bias mitigation methods in Table \ref{tab:mitigation} together with the respective publications.

\subsection{Debiasing Using Data Manipulation}
\label{sec:debias-corpus}

Debiasing using data manipulation commonly refers to counterfactual data augmentation, gender tagging, adding context, and balanced fine-tuning. Below we describe these approaches in detail.

\subsubsection{Data Augmentation}
\label{sec:debias-cda}

Many concerns have been posed regarding modern NLP systems having been trained on potentially biased datasets, as as these biases can be perpetuated to downstream tasks and eventually society in the form of allocational harms \citep{hovy2021sources}. Therefore, \citet{costa-jussa-de-jorge-2020-fine} claim that developing methods trained on balanced data is a first step to eliminating representational harms.

In order to attenuate the impact of gender bias from the dataset used, \citet{zhao-etal-2018-gender} propose a rule-based approach to generate an auxiliary dataset where all-male entities are replaced by female entities (and vice-versa) and suggest to train methods on the union of the original and augmented dataset. Thus, both male and female genders are equally represented in the dataset. For instance, a sentence \textit{My son plays with a car.} would be transformed into \textit{My daughter plays with a car}. Therefore, to apply this method, a list of gendered pairs (such as \textit{son--daughter}) is required. 
Similarly, \citet{emami-etal-2019-knowref} propose to extend a training set for coreference resolution by switching every entity pair. 
A method for debiasing gender-inflected languages is demonstrated in \citet{zmigrod-etal-2019-counterfactual}, where sentences are reinflected from masculine to feminine (and vice-versa) in a counterfactual data augmentation (CDA) scheme. Since this method analyses each word separately, it is not applicable to more complex sentences involving coreference resolution. However, it introduces a feasible debiasing approach for languages beyond English. \citet{hall-maudslay-etal-2019-name} develop a name-based version of CDA, in which the gender of words denoting persons in a training corpus are swapped probabilistically in order to counterbalance bias.

Due to its simple implementation, counterfactual data augmentation has been widely applied to mitigate gender bias. Since the model observes the same scenario in the doubled (for binary gender) sentences, it can learn to abstract away from the entities to the context \citep{emami-etal-2019-knowref}. This method has shown encouraging results in mitigating bias in contextualised word representations such as ELMo and monolingual BERT \citep{zhao-etal-2019-gender, bartl-etal-2020-unmasking, de-vassimon-manela-etal-2021-stereotype,sen-etal-2021-counterfactually}, and for hate speech detection \citep{park-etal-2018-reducing}.
Nonetheless, collecting annotated lists for gender-specific pairs can be expensive, and the method essentially doubles the size of the training data. To this end, \citet{de-vassimon-manela-etal-2021-stereotype} compare fine-tuning contextualised word representation on augmented and un-augmented datasets and show that fine-tuning solely on an augmented corpus successfully decreases gender bias. 


Another method of gender bias mitigation via data augmentation is presented in \citet{stanovsky-etal-2019-evaluating} who suggest a simple approach of ``fighting bias with bias'' and add stereotypical adjectives to describe entities of the respective gender, e.g., \textit{``The pretty doctor asked the nurse to help her in the procedure.''}. However impractical this method is, relying on accurate coreference resolution, it has shown to reduce bias in the tested languages.

\subsubsection{Gender Tagging}
Another stream of work has concentrated on incorporating external or internal gender information during training. This method has been widely employed in debiasing neural machine translation models to mitigate the issue of male default. \citet{moryossef-etal-2019-filling} append a short phrase at inference time which acts as an indicator for the speaker's gender, e.g., \textit{``She said:''}, while similarly, \citet{vanmassenhove-etal-2018-getting} use sentence-level annotations. In order to extend the mitigation method to be applicable to sentences with more than one gendered entity, \citet{stafanovics-etal-2020-mitigating} utilise token-level annotations for the subject's grammatical gender. 
\citet{habash-etal-2019-automatic} propose a post-processing method that is an intersection of gender tagging and CDA and test it on Arabic. 
In gender-aware debiasing, a gender-blind system is being turned into a gender-aware one by identifying gender-specific phrases in the system's output and subsequently offering alternative reinflections.
In the domain of machine translation, \citet{saunders-etal-2020-neural} propose an approach based on fine-tuning a model on a small, artificial dataset of sentences with gender inflection tags which are then replaced by placeholders. However, the results of this scheme are ambiguous, and this method is not well suited for translating sentences with multiple entities.

Methods relying on gender tagging are a flexible tool for controlling for bias.  
However, we note that these methods do not inherently remove gender bias from the system \citep{cho-etal-2019-measuring}. Additionally, gender tagging requires meta-information on the gender of the speaker, which is often either expensive or unavailable. 

\subsubsection{Adding Context}

Alongside including the speaker's information as in the above examples, \citet{basta-jussa-2020-mitigating} concatenate the previous sentence from a corpus to increase the context. Using the additional information only in the decoder part of the Transformer architecture ultimately reduces training parameters, simplifies the model, and requires no further information for training or inference. \citet{basta-jussa-2020-mitigating} show that this method improves the performance of machine translation with coreference resolution tasks. However, \citet{savoldi2021gender} note that this improvement might not be due to the added gender context, but for instance, a regularisation effect.


\subsubsection{Balanced Fine-Tuning}

Balanced fine-tuning incorporates transfer learning from a less biased dataset. In the first step, a model is trained on a large, unbiased dataset for the same or similar downstream task and is then fine-tuned on a target dataset which is more biased \citep{park-etal-2018-reducing}. Such a training regime obviates potential over-fitting to a biased dataset. 
This method suffers from a severe limitation, namely assuming an existence of an unbiased dataset in its initial step, which is usually infeasible to obtain and thus, not applicable in real-life applications.
On the other hand, \citet{saunders-byrne-2020-reducing} consider gender bias in machine translation as a domain adaptation task and use a handcrafted gender-balanced dataset 
together with a lattice re-scoring module to mitigate the consequences of initial training on unbalanced data. \citet{saunders-byrne-2020-reducing} consider three aspects of the adaptation problem: creating less biased adaptation data, parameter adaptation using this data, and inference with the debiased models produced by adaptation. However, the need for a gender-balanced dataset for a specific domain might be a drawback of this approach.  
\citet{costa-jussa-de-jorge-2020-fine} use an inverse approach and train their model on a larger corpus and fine-tune it with a gender-balanced corpus showing that their approach successfully mitigates gender bias and increases performance quality even if the balanced dataset is coming from a different domain. However, \citet{savoldi2021gender} note that this approach does not account for the qualitative differences in how men and women are portrayed \citep{savoldi2021gender}.

\subsection{Debiasing by Adjusting Algorithms}
\label{sec:debias-algo}

Instead of manipulating the underlying data, a number of gender debiasing methods have been implemented to approach the issue via algorithm adjustment. Such techniques can be categorised into the following groups: projection-based debiasing, constraining models' predictions, applying adversarial learning approaches, and other.

\subsubsection{Projection-Based Debiasing}

To the best of our knowledge, \citet{schmidt2015rejecting} propose the first word embedding debiasing algorithm and remove multiple gender dimensions from word vectors. In parallel, instead of completely removing gender information, \citet{bolukbasi2016man} suggest shifting word embeddings to be equally male and female in terms of their vector direction. For instance, a debiased embeddings for \textit{grandmother} and \textit{grandfather} will be equally close to \textit{babysit} without neglecting the analogy to gender. More formally, \citet{bolukbasi2016man} propose two debiasing methods, hard- and soft-debiasing. The first step for both of them consists of identifying a list of gender-neutral words and a direction of the embedding that captures the bias.
\textbf{Hard-debiasing} (or `Neutralise and Equalise method') ensures that gender-neutral words are zero in the gender subspace and equalises sets of words outside the subspace and thereby enforces the property that any neutral word is equidistant to all words in each equality set (a set of words which differ only in the gender component). For instance, if (grandmother, grandfather) and (guy, gal) were two equality sets, then after equalisation, `babysit' would be equidistant to grandmother and grandfather and also to gal and guy, but closer to the grandparents and further from the gal and guy. This approach is suitable for applications where one does not want any such pair to display any bias with respect to neutral words.
The disadvantage of equalising sets of words outside the subspace is that it removes particular distinctions that are valuable in specific applications. For instance, one may wish a language model to assign a higher probability to the phrase to `grandfather a regulation' since it is an idiom, unlike `grandmother a regulation'. The \textbf{soft-debiasing} algorithm reduces differences between these sets while maintaining as much similarity to the original embedding as possible, with a parameter that controls for this trade-off. In particular, soft-debiasing applies a linear transformation that seeks to preserve pairwise inner products between all the word vectors while minimising the projection of the gender-neutral words onto the gender subspace.

Both hard- and soft-debiasing approaches have been applied in research to word embeddings and language models. \citet{bordia-bowman-2019-identifying} validate the soft-debiasing approach to mitigate bias in LSTM based word-level language models. \citet{park-etal-2018-reducing} compare hard-debiasing method to other methods in the context of abusive language detection. \citet{sahlgren-olsson-2019-gender} apply hard-debiasing to Swedish word embeddings and show that this method does not have the desired effect when tested on selected downstream tasks.
\citet{sahlgren-olsson-2019-gender} argue that these unsatisfactory results are due to including person names in their training process.
Interestingly, \citet{ethayarajh-etal-2019-understanding} show that debiasing word embeddings post hoc using subspace projection is, under certain conditions, equivalent to training on an unbiased corpus.
Similarly to \citet{bolukbasi2016man}, \citet{karve-etal-2019-conceptor, sedoc-ungar-2019-role} aim to identify the bias subspace in word embeddings using a set of target words and a \textbf{debiasing conceptor}, a mathematical representation of subspaces that can be operated on and composed by logic-based manipulations. 

However, these methods strongly rely on the pre-defined lists of gender-neutral words \citet{sedoc-ungar-2019-role}. 
Moreover, \citet{zhao-etal-2018-learning} prove that an error in identifying gender-neutral words affects the performance of the downstream model. 
\citet{bordia-bowman-2019-identifying, zhao-etal-2018-learning} notice a trade-off between perplexity and gender bias as in an unbiased model, male and female words are predicted with an equal probability. This can be undesirable in domains such as social science and medicine. 
While \citet{gonen-goldberg-2019-lipstick} claim that debiasing is primarily superficial since a lot of the supposedly removed bias can still be recovered due to the geometry of the word representation of the gender neutralised words, \citet{prost-etal-2019-debiasing} show that soft-debiasing can even increase the bias of a downstream classifier by removing noise from gender-neutral words and ultimately providing a less noisy communication channel for gender clues.

Recently, \citet{liang-etal-2020-monolingual} use DensRay \citep{dufter-schutze-2019-analytical}, an interpretable method for identifying the embedding subspace using projections and then evaluate gender bias in masked language models by comparing the difference in the log-likelihood between male and female pronouns in a template \textit{``[MASK] is a/an [occupation].''}. However, this method heavily relies on a list of occupations and a simple template. 
Further, \citet{Dev_Li_Phillips_Srikumar_2020} employ an orthogonal projection to gender direction \citep{pmlr-v89-dev19a} to debias contextualised embeddings and test it on a NLI task with gender-biased hypothesis pairs.
However, this method can only be applied to the model's non-contextualised layers. 
\citet{kaneko-bollegala-2021-debiasing} obviate this limitation in a fine-tuning setting. Their method applies an orthogonal projection only in the hidden layers and proves to outperform \citet{Dev_Li_Phillips_Srikumar_2020}. Additionally, this method is independent of model architectures or their pre-training method. However, this approach requires a list of attribute words (e.g., she, man, her) and target words (e.g., occupations) to extract relevant sentences from the corpus, making their method highly reliant on this list. 

\subsubsection{Constraining Output}

A simple approach to debiasing algorithms is to constrain model output post-hoc. To this end, \citet{zhao-etal-2017-men} propose a debiasing techique that constrains model predictions to follow a distribution from a training corpus, e.g., the ratio of male and female pronouns. Thus, this method is highly dependent on the gender balance and bias in the underlying data. 

In the field of language generation, \citet{ma-etal-2020-powertransformer} introduce \textit{controllable debiasing} as an unsupervised text revision task that aims to correct the implicit bias against or towards a specific character portrayed in a language model generated text. For this purpose, they create an encoder-decoder model that rewrites a text to portray females as more agent (in terms of \citet{sap-etal-2017-connotation}'s connotation frames). However, their approach relies strongly on an external corpus of paraphrases. 

\subsubsection{Adversarial Learning}

Another strain of work has employed adversarial learning as a debiasing method. \citet{li-etal-2018-towards} propose a method for removing model biases by explicitly protecting demographic information (such as gender) during model training. However, \citet{elazar2018adversarial} claim that word representations preserve traces of the protected attributes and recommend external verification of the method.
Similarly, \citet{zhang2018mitigating} apply adversarial learning by including gender as a protected variable and having the generator learn with respect to it.
In general, the objective of such a model is to maximise the predictor's ability to predict a variable of interest while fooling the adversary to predict the protected attribute. However, in general, adversarial learning is often an unstable method and can only be used when gender is a protected attribute rather than a variable of interest. 




\subsubsection{Other}

Several other methods have been tested to mitigate gender bias in NLP methods. 

Alongside projection-based methods for debiasing word embeddings, another approach to debiasing word embeddings has aimed to learn their gender-neutralised variant. In particular, \citet{zhao-etal-2018-learning} propose to train word embeddings such that protected attributes are neutralised in some of the dimensions, resulting in gender-neutral word representations. Restricting the information of protected attributes in certain dimensions enables its removal from an embedding. Additionally, other than the method presented in \citet{bolukbasi2016man} gender-neutral words are learned jointly in the training process instead of being manually created. However, \citet{sun-etal-2019-mitigating} note that it is unclear if gender-neutralised word embeddings are applicable to languages with grammatical genders.

Adjusting the loss function has proven to be another viable method for gender bias mitigation. In particular, \citet{qian-etal-2019-reducing} introduces a new term to the loss function, which attempts to equalise the probabilities of male and female words (based on a pre-defined list) in the output and evaluate it on a text generation task. We see two main limitations of this approach. First, it relies on a straightforward definition of bias (\textit{i.e.}, an equal number of gender mentions). Second, as with many other methods, it requires a list of gender pairs, a limitation we discuss above.  

Gender-preserving debiasing has been introduced to mitigate gender bias, accounting that not all gender associations are stereotypical. \citet{kaneko-bollegala-2019-gender} split a given vocabulary into four mutually exclusive sets of word categories: words that are female-biased but non-discriminative, male-biased but non-discriminative, gender-neutral words, and words perpetuating stereotypes.
\citet{kaneko-bollegala-2019-gender} learn word embeddings that preserve the information for the gendered but non-stereotypical words protects the neutrality of the gender-neutral words while removing the gender-related biases from stereotypical words. The embedding is learnt using an encoder in a denoising autoencoder, while the decoder is trained to reconstruct the original word embeddings from the debiased embeddings. However, creating a word list with the above-mentioned categories of words is time-consuming, and word categorisation might not be straightforward. 

\citet{jin-etal-2021-transferability} investigate incorporating bias mitigation into the model's objective. First, an upstream model is fine-tuned with a bias mitigation objective which consists of a text encoder and a classifier head. Subsequently, the encoder from the upstream model, jointly with the new classification layer, are again fine-tuned on a downstream task. \citet{jin-etal-2021-transferability} note that upstream bias mitigation, while less effective, is more efficient than direct bias mitigation methods without fine-tuning. However, it requires a tailored evaluation for the downstream task.  






\section{Discussion}

After presenting probing datasets, formal definitions, detection, and mitigation methods, we next present the main findings we make throughout this survey. 
We find that existing research on gender bias has four main limitations and discuss them in the following.

\paragraph{Gender in NLP}

It is not uncommon for studies about gender to be reported without any explanation of how gender labels are ascribed, 
and the ones that do, explain the imputation of gender categories in a debatable way \citep{larson-2017-gender}. Using gender as a variable in NLP is an ethical issue, thus unreflectively assigning gender category labels may violate ethical frameworks that demand transparency and accountability from researchers \citep{larson-2017-gender}. Therefore, it is crucial to ask how researchers can use NLP tools to investigate the relationship between gender and text meaningfully, yet without harmful stereotypes \citet{koolen-van-cranenburgh-2017-stereotypes}. To obviate this risk, \citet{larson-2017-gender} suggest formulating research questions with explicit definitions of \textit{gender}, avoiding using gender as a variable unless it is necessary. 
Not being explicit about the ascription of the category of gender as a variable to participants brings into question the internal and external validity of research findings because it makes it difficult to near-impossible for other scholars to reproduce, test, or extend study findings \citep{larson-2017-gender}. 


We find that researchers often decide to define gender in their study as binary. However, making this assumption is an oversimplification of gender complexity and can perpetuate harms to non-binary people \citep{fast2016shirtless, behm2008}. We encourage researchers to define gender in a  transparent and inclusive manner, to expand corpora with inclusive pronouns, and evaluate models on non-binary pronouns as well to mitigate these harms. So far models' performance on downstream tasks has been consistently lower for non-binary pronouns compared to the binary pronouns \citep{Sun2021TheyTT,cao-daume-iii-2020-toward}.




\paragraph{Monolingual focus}

Gender bias is grounded in societal and cultural views on gender, and thus, its expressions vary across languages. Expanding research to languages beyond English and including data from outside of the Anglosphere would lead to gaining a broader view on gender bias in societies which we strongly encourage. However, most prior research on gender bias has been monolingual, focusing predominantly on English or a small number of other high-resource languages such as Chinese \citep{liang-etal-2020-monolingual} and Spanish \citep{zhao-etal-2020-gender} with the notable exception of a broader multilingual analysis of gender bias in machine translation \citep{prates2019assessing} and language models \citep{stanczak2021quantifying}.  

\paragraph{Need for formal testing}
Most of the papers that have focused on detecting gender bias in natural language, methods, or downstream tasks, have seen bias detection as a goal in itself or a means of analysing the nature of bias in domains of their interest.
Widely acknowledged models that have led in recent years to significant gains on many NLP tasks have not included any study of bias alongside the publication \citep{conneau-etal-2020-unsupervised,devlin-etal-2019-bert,peters-etal-2018-deep,Radford2019LanguageMA}.
In general, these methods are tested for biases only post-hoc when already being deployed in real-life applications, potentially posing harm to different social groups \citep{mitchell2019}. Since these models were probed for gender bias only after their release, they might have already caused societal harms. We find that bias detection should be included in the model development pipeline at early stages and see enforcing this change as a primary challenge. 
The way to ensure that researchers abide by ethical principles is to hold them accountable when research projects are planned, \textit{i.e.}, requiring project proposals and publications to include ethical considerations and, later, during the peer review process.

\paragraph{Limited definitions}
However, to introduce formal testing comprehensive and multi-faceted bias measures are required. 
We find that similarly to research within societal biases \citet{blodgett2020language}, work on gender bias in particular, suffers from incoherence in usage of evaluation metrics. Most of the publications on gender bias consider only one way of defining bias and do not engage enough parallel research to combine these methods. Gender bias can be expressed in language in many nuanced ways which poses stating a comprehensive definition as one of the main challenges in this research field. Finally, we strongly encourage developing standard evaluation benchmarks and tests to enhance comparability.

\section{Conclusion}

In this paper, we present a comprehensive survey of \XX papers on gender bias in natural language and NLP methods published since gender bias has been studied in NLP. We find four major limitations in the existing research and see overcoming these limitations as crucial for further development of this field.  

First, most research lacks transparent and inclusive gender and gender bias definitions. Gender is mainly treated as a binary variable which disagrees with social science position. Next, the majority of the work disregards low-resource languages, concentrating solely on English and other high-resource languages such as Spanish and Chinese, which imposes a strongly restricted view on the nature of gender bias in NLP. 
Moreover, despite a myriad of papers on gender bias in NLP methods, most of the newly developed algorithms do not test their models for bias and disregard possible ethical considerations of their work. This leads to deployment of models that lead to potential societal harms. 
Finally, we find that the methodology used in this research field is fundamentally flawed, covering only limited aspects of gender bias and lacking baselines for evaluation and testing pipelines. 


\bibliographystyle{ACM-Reference-Format}
\bibliography{gender-base}

\appendix

\end{document}